\documentclass[10pt, conference]{IEEEtran}
\usepackage{graphicx} % for pdf, bitmapped graphics files
\usepackage{url}
\usepackage{algorithm}
\usepackage{algpseudocode} 
\usepackage{siunitx}
\usepackage{bookmark}
\usepackage{paralist}
\usepackage{subfiles}
\usepackage[]{caption}
\usepackage{threeparttable}
\usepackage{hyperref}

\begin{document}
\title{ROS-X-Habitat:\\ Bridging the ROS Ecosystem with Embodied AI}


\author{\IEEEauthorblockN{Guanxiong Chen, Haoyu Yang, Ian M. Mitchell}
\IEEEauthorblockA{Department of Computer Science\\
The University of British Columbia\\
Vancouver, BC\\
Email: \{gxchen, yhymason, mitchell\}@cs.ubc.ca}
%\and
%\IEEEauthorblockN{Ian M. Mitchell}
%\IEEEauthorblockA{Department of Computer Science\\
%The University of British Columbia\\
%Vancouver, BC\\
%Email: mitchell@cs.ubc.ca}
}

%Telephone: (800) 555--1212, Fax: (888) 555--1212}


\maketitle

\begin{abstract}
We introduce ROS-X-Habitat, a software interface that bridges the AI Habitat platform for embodied learning-based agents with other robotics resources via ROS. This interface not only offers standardized communication protocols between embodied agents and simulators, but also enables physically and photorealistic simulation that benefits the training and/or testing of vision-based embodied agents. With this interface, roboticists can evaluate their own Habitat RL agents in another ROS-based simulator or use Habitat Sim v2 as the test bed for their own robotic algorithms. Through in silico experiments, we demonstrate that ROS-X-Habitat has minimal impact on the navigation performance and simulation speed of a Habitat RGBD agent; that a standard set of ROS mapping, planning and navigation tools can run in Habitat Sim v2; and that a Habitat agent can run in the standard ROS simulator Gazebo.
\end{abstract}


% \fi
\IEEEpeerreviewmaketitle

\section{Introduction}
Since the earliest days of robotics, researchers have sought to build embodied agents to perform a variety of jobs, such as assistive tasks in factories \cite{rl-for-manufacturing} or wildfire surveillance \cite{rl-for-wildfire-surv}. Following tremendous advancements in deep learning and convolutional neural networks over the past decade, researchers have been able to develop learning-based embodied agents that interact with the real world on the basis of visual observations \cite{duan2022}. Software platforms such as OpenAI Gym \cite{openai-gym}, Unity ML-Agents Toolkit \cite{unity}, Nvidia Isaac Sim \cite{isaac} and AI Habitat \cite{savva2019habitat} have emerged to address the community's need for training and evaluating learning-based embodied visual agents end-to-end. Our research group was particularly intrigued by the AI Habitat platform, because this open source platform offers developers not only flexibility to tailor the software to their own needs, but also a high-performance photorealistic simulator recently enhanced with the Bullet physics engine \cite{bullet}, as well as access to a sizeable library of visually-rich scanned 3D environments with which to bridge the \emph{sim2real} performance gap for visual agents \cite{garcia-garcia2018robotrix}.

Even though these platforms allow roboticists to reuse existing learning algorithms and train visual agents in simulators with ease, there is a critical step to using them for embodied agents which is only partially addressed: Connecting the trained agent with a robot. After training a learning-based agent in simulation one would like to take advantage of the extensive set of tools and knowledge from the robotics community to make it easy to embody that agent. One particularly popular tool from the robotics community is ROS, a robotics-focused middleware platform with extensive support for classical robotic mapping, planning and control algorithms, as well as drivers for a wide variety of computing, sensing and actuation hardware.  But ROS' support for direct interfacing with RL agents is limited, and Gazebo---the standard simulation environment used for ROS systems---suffers from two significant shortcomings when applied to this task:
%\begin{inparaenum}[(i)]
\begin{inparaenum}[(i)]
    \item It lacks the level of photorealism required to allow vision-based agents trained with synthetic inputs to perform well in real world scenarios \cite{garcia-garcia2018robotrix, gaidon2016virtual};
    \item It cannot match the simulation speed of tools specifically designed to train large-scale reinforcement learning (RL) agents \cite{parallel-gym-gazebo}.
\end{inparaenum}
%\end{inparaenum}

%Meanwhile, robotics researchers are still developing classical embodied agents that follow the ``sense-plan-act'' approach \cite{sense-plan-act}. Compared with learning-based agents, classical agents tend to be less data and compute intensive, and  Many developers build and test classical agents for indoor tasks within the ROS ecosystem. For navigation tasks, packages such as \verb|hector_mapping| \cite{hector} and \verb|move_base| \cite{movebase} allow users to easily map an environment and set up a planner. Simulation tools such as Gazebo that work under ROS offer physically realistic simulation. The vast amount of hardware drivers from ROS packages allows users to easily test an agent in the real world. But ROS also has its own problems: 1) little support for building learning-based agents, and 2) limited choices of simulation tools - while Gazebo is still the canonical choice for people working under ROS, and it provides users freedom to author their own 3D assets and scenes, the simulator lacks photorealism and simulation speed required to train RL agents on a large scale \cite{parallel-gym-gazebo}.

\begin{figure}
    \centering
    \includegraphics[width=.45\textwidth]{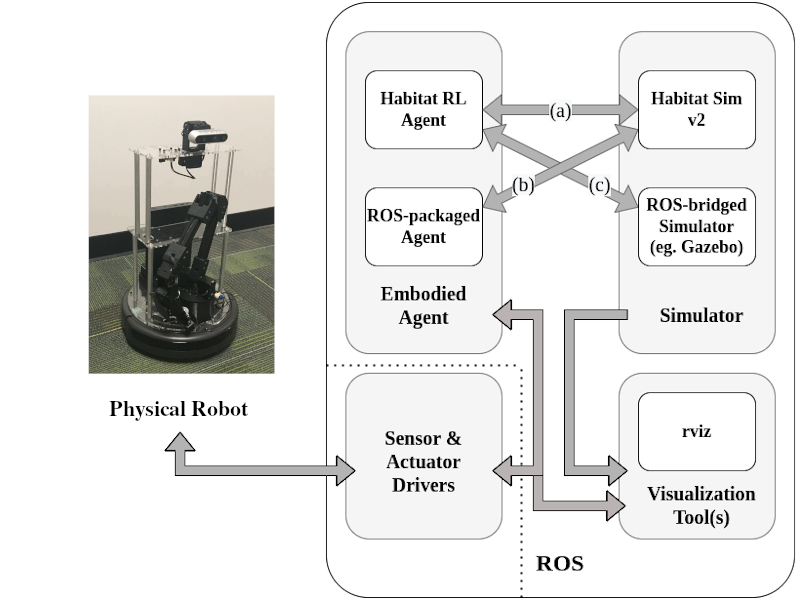}
    \caption{High-level overview of \texttt{ros\_x\_habitat}'s architecture.
    In this paper we demonstrate the system's operation under modes (a), (b) and (c) to interface ROS or Habitat agents with a ROS-based simulator or Habitat Sim v2.
    %Each operating mode is detailed in Section \ref{sec:arch}. The interface also provides visualization tools under ROS such as \texttt{rviz} access to sensory observations from Habitat Sim v2.
    Although not demonstrated in the paper, the modular design of \texttt{ros\_x\_habitat} allows easy embodiment of a Habitat visual navigation agent on a physical robot (such as the LoCoBot \cite{locobot}).}
    \label{fig:high_level_arch}
\end{figure}

In order to take advantage of the strengths and overcome the weaknesses of these two independent sets of tools, in this paper we present ROS-X-Habitat (``ROS-Cross-Habitat''), an interface that bridges the AI Habitat training platform with the ROS ecosystem. Figure \ref{fig:high_level_arch} shows a simplified view of the interface's architecture. The interface makes the following contributions to the robotics community:
\begin{inparaenum}[(i)]
    \item It allows embodied agents to leverage Habitat Sim v2's physics-based simulation capability through the interface. We demonstrate that the performance of a Habitat agent trained in a simulation environment without physics degrades only slightly when physics is turned on via the interface.  This result suggests that high throughput training \emph{without} the overhead of ROS and physics can produce agents that are still effective once these additional layers are added.
    \item It allows AI Habitat's vision-based RL agents to be evaluated in ROS, so RL agent developers can take advantage of ROS's rich set of tools and community support, as well as Gazebo's ability to let users author 3D assets and customary testing scenes.  Although not demonstrated here, this bridge to ROS dramatically shortens the path to embodying the Habitat agents in a physical robot.
    \item It allows other embodied agents implemented in ROS packages---for example, classical navigation algorithms such as \cite{dwa, movebase}---access to the rich 3D environments available in Habitat Sim v2.  Consequently, algorithms from other sources can be evaluated in these photorealistic environments.
\end{inparaenum}
Additional details and features of the project can be found in~\cite{chen2021ros, Yang_2022}.  The source code of the interface is available at \url{https://github.com/ericchen321/ros_x_habitat}.

\section{Related Work}
Based on the objectives of our work, we review previous work in three relevant areas of research: 
\begin{inparaenum}[(i)]
\item embodied agents, \item robotic middleware, \item simulators and datasets for training and evaluating learning-based embodied agents or for evaluating classical embodied agents in general.
\end{inparaenum}
%From the existing work, we show the robotics community's dire need to
%\begin{inparaenum}
%    \item bridge learning-based embodied agents with Gazebo or other ROS-bridged simulators;
%    \item bridge classical ROS-based planners with state-of-the-art robotic simulators that offer greater level of photorealism;
%    \item evaluate embodied agents' performance in a continuous action space to leverage simulators' physics simulation capability. 
%\end{inparaenum}

\subsection{Embodied Agents}
We consider two categories of embodied agents commonly used to complete navigation tasks: Classical robotics approaches and learning-based.

Most commonly deployed classical embodied agents do navigation in two phases: Construct a map of the environment using, for example, a SLAM algorithm (such as \cite{rtabmap, hector, orb-slam2}), and then use the map to plan out a path to the goal position (such as \cite{dwa,movebase}).  While many packages that follow this approach are available in ROS, most recent photorealistic, physics-capable simulators (see Section~\ref{sec:related-sim}) do not interface to ROS and so evaluating classical agents in those simulators is difficult. So-called ``end-to-end'' learning-based agents use a neural network to produce a sequence of actions directly from visual observations and localization data without relying on prior maps \cite{savva2019habitat,tai2017virtual,datta2020integrating}. Another popular approach is to combine learning-based agents with classical mapping-then-planning \cite{Brunner_Richter_Wang_Wattenhofer_2018, neural-slam, nikdel2020recognizing}. But none of these frameworks connect directly to ROS.

Although not fundamental to their designs, a common distinction between classical and visual learning-based agents is that the former operate in a continuous action space while the latter (especially RL agents) are often trained to produce discrete actions. Continuous action spaces are more representative of how physical robots actuate~\cite{masson2016reinforcement}, but training for these spaces has higher computational cost~\cite{discrete-drl}. As an example, Habitat's default PointGoal navigation agents have an action space consisting of four actions~\cite{savva2019habitat}: \verb|move_forward|, \verb|turn_left|, \verb|turn_right|, and \verb|stop|. To simulate these actions, Habitat Sim teleports the robot from one state to another without taking account of interactions between the agent and other objects at intermediate states.  It is certainly possible to map from discrete to continuous actions, but it is not clear a priori that embodying a discrete agent into a continuous action space in this manner will produce good results.

\subsection{Simulators and Datasets} \label{sec:related-sim}
The exploding interest in RL agents, and particularly vision-based RL agents, has put a premium on simulation speed and photorealism.  Gazebo \cite{gazebo} has been the canonical choice for simulation in the ROS community; it benefits from extensive community support and a huge set of community-shared prebuilt assets, but lacks photorealism and high simulation speed. Unity is a powerful game engine, and the Unity ML-Agents Toolkit available to researchers provides simulation environments suitable for embodied agents \cite{unity}, but the toolkit does not provide photorealistic simulation spaces by default and lacks compatibility with off-the-shelf 3D datasets such as Replica \cite{straub2019replica} or Matterport3D \cite{chang2017matterport3d}. Isaac Sim \cite{isaac} shows great promise in terms of configurability and photorealism, but is not currently open-source. The recently released Sapien \cite{xiang2020sapien} platform offers a photorealistic and physics-rich simulated environment, but currently provides limited support for tasks other than motion planning.

In this paper we explore the use of Habitat Sim v2 from the AI Habitat platform \cite{szot2021habitat} for several reasons:
\begin{inparaenum}[(i)]
    \item Extremely high speed simulation. This feature is particularly useful for RL agents, since agent performance may continue to improve even after many millions of training steps.
    \item Photorealistic rendering of scanned spaces. Habitat Sim v2 can render photorealistic scenes (including depth maps) from Habitat's native datasets such as Replica \cite{straub2019replica} and ReplicaCAD \cite{szot2021habitat}. The Habitat framework's modular design~\cite{savva2019habitat, szot2021habitat} offers a seamless interface to many prior photorealistic 3D scene datasets, including Matterport3D \cite{chang2017matterport3d} and Gibson \cite{xia2018gibson}.  For this paper we use the Matterport3D test set, since it is the only publicly available dataset for which Habitat navigation results have been reported against which we can compare (albeit for v1 agents) \cite{savva2019habitat}.
    \item Simulation of many different tasks.  We focus here on PointGoal navigation~\cite{savva2019habitat} from Habitat v1, but object picking has also been demonstrated~\cite{szot2021habitat}.
\end{inparaenum}

\subsection{Robotics Middleware}
Robotics middleware~\cite{middleware} is an abstraction layer that resides between the robotics software and the operating system (which is itself abstracting the underlying hardware). Middleware provides standardized APIs to sensors, actuators, and communication; design modularity; and portability. 

A variety of robotics middleware systems have been developed (for example, see \cite{kramer07}), each with its own strengths and limitations. For example, while Microsoft RDS supports multiple programming languages, it runs only on the Windows OS\cite{survey}. OROCOS offers its own optimized runtime environment for real-time applications, but it does not have a graphical environment for drag-and-drop development nor a simulation environment \cite{survey}. ROS (Robot Operating System) is a free, open-source and popular robotics middleware introduced in 2007 \cite{ros}. Among its features: \begin{inparaenum}[(i)]
    \item It promotes modular, robust and potentially distributed designs;
    % by breaking implementations into communicating nodes and services that run independently, 
    \item The huge user community has generated thousands of ready-to-use packages, including drivers for all common robotics hardware~\cite{ros, survey-of-dev-framework-for-rob, comparison}. 
\end{inparaenum}

Given the popularity of ROS, it should come as no surprise that others have sought to build an interface between an embodied AI platform and the ROS ecosystem. PathBench is a unified interface that allows a multitude of path-planning algorithms to be evaluated under ROS and Gazebo \cite{toma2021pathbench}. Zamora et al. \cite{DBLP:journals/corr/ZamoraLVC16} used ROS as a bridge between OpenAI Gym and Gazebo, but did not consider other simulation environments which could provide photorealism. The Habitat-PyRobot Bridge (HaPy) \cite{sim2real} is particularly related to our work: it provides integration between Habitat agents and PyRobot, which is itself an interface on top of ROS to control different robots \cite{pyrobot}. A key distinction between our work and HaPy-PyRobot is a design decision about how much the user is exposed to ROS.  PyRobot abstracts ROS components away from top-level APIs \cite{pyrobot}, thus requiring less familiarity with ROS from its users.  ROS-X-Habitat exposes ROS components to a greater extent, thus requiring more familiarity with ROS but allowing users greater flexibility to take advantage of the full ROS ecosystem.  Also, HaPy does not currently support physics-based simulation from Habitat Sim.

% \begin{inparaenum}[(i)]
%     \item While HaPy-PyRobot abstracts ROS components away from top-level APIs \cite{pyrobot} because it targets at roboticists not familiar with ROS, ROS-X-Habitat exposes ROS components to a greater extent for easier integration with other ROS tools and thus offers good extensibility under the ROS framework, because we target at users with basic knowledge in ROS;
%     \item Many of PyRobot's utilities can only interface with a limited range of Habitat's functionalities because their APIs were built to work with simulation and training platforms other than Habitat as well. Our work allowing users to easily interface with a broader range of functionalities offered by Habitat.
% \end{inparaenum}

\section{System Design}
\subsection{Requirements} \label{sec:req}
The interface should allow users to:
\begin{compactenum}
    \item \textbf{Leverage Habitat Sim v2's full physics-based simulation capability to provide a more realistic environment for embodied agents.}  This step is non-trivial because (at the current time) Habitat Sim v2 cannot simulate agents which produce discrete actions with full dynamic physics.  In order to do so we have implemented API calls in the interface to enable physics-based simulation. Section~\ref{sec:hab-hab} demonstrates success on this requirement for a learned Habitat RGBD navigation agent, and Section~\ref{sec:ros-hab} demonstrates success for a classical ROS planner.
    \item \textbf{Evaluate an agent inside a simulator through the interface with minimal performance impact.} Section~\ref{sec:hab-hab} reports a number of experiments designed to gauge the effect of the ROS middleware on the performance of the Habitat platform components.
    \item \textbf{Deploy ROS-based planners to navigate within Habitat Sim v2, thus allowing researchers to test planners on a huge variety of photorealistic 3D scenes.}  Section~\ref{sec:ros-hab} demonstrates that this combination is feasible using the Matterport 3D dataset.
    \item \textbf{Evaluate Habitat agents in other simulation environments with ROS bridges or on physical hardware.} Section~\ref{sec:hab-gazebo} demonstrates this capability in Gazebo, and once we know that this combination works the jump to other simulators or a real robot is much less daunting.
    \item \textbf{Extend or customize its functionality for bridging with other tools from both the ROS and Habitat frameworks.} In Section \ref{sec:arch} we show the modular design of ROS-X-Habitat.  The exposure of ROS nodes, topics and services allows users to interface to other ROS tools (such as RViz) or Habitat's visualization modules.
\end{compactenum}

\subsection{Architecture} \label{sec:arch}

\begin{figure*}[ht]
    \centering
    \includegraphics[width=.9\textwidth]{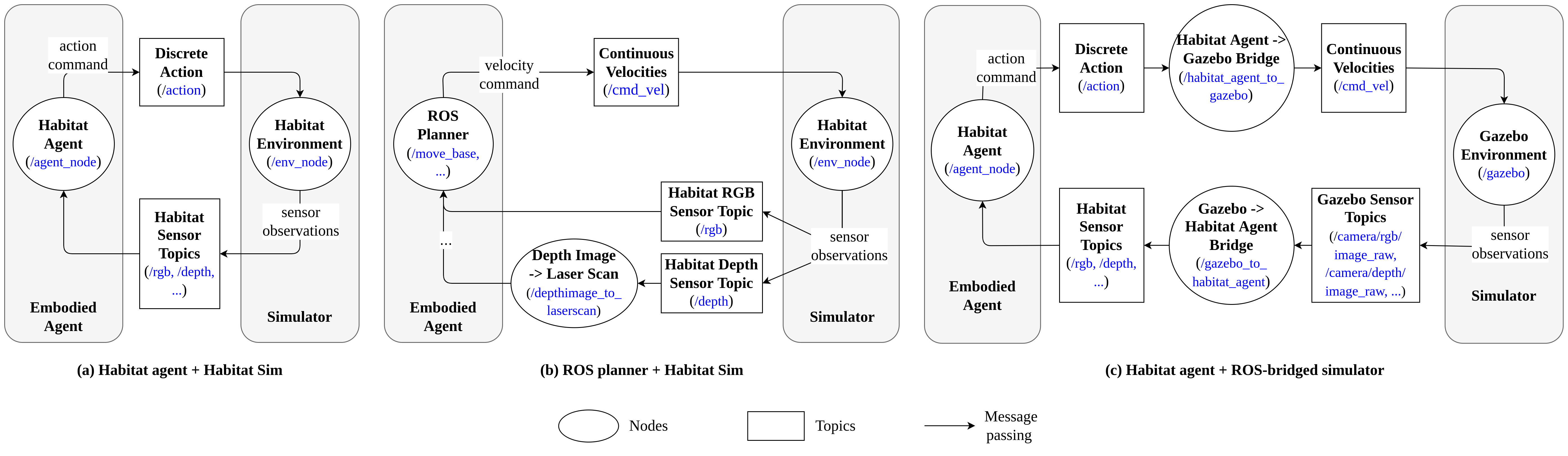}
    \caption{The \texttt{ros\_x\_habitat} architecture under each operating mode. (a) Simulating a Habitat RL agent in Habitat Sim v2. If physics is turned on, each discrete action would be converted to a sequence of velocity commands in the node encapsulating the Habitat environment (\texttt{env\_node}) following Algorithm \ref{alg:action_to_vel}. (b) Simulating a ROS-based planner in a physics-enabled and photorealistic Habitat Sim v2 environment. (c) Simulating a Habitat RL agent in a Gazebo environment.  Note that for simplicity, we omitted nodes, topics and services not being the gist of the interface's operation---for example, map topics and nodes under operating mode (b).}
    \label{fig:architecture}
\end{figure*}

We designed ROS-X-Habitat as a ROS Noetic package (\verb|ros_x_habitat|), written in Python and consisting of a collection of ROS nodes, topics and services. Following the five requirements we established, we defined three operating modes for the interface, as illustrated in Figure \ref{fig:architecture}.

\begin{compactenum}[(a)]
    \item \textbf{A Habitat agent navigates in a Habitat Sim-rendered scene (with or without physics) through ROS.} Note that this is not an expected use case for \verb|ros_x_habitat|; after all, running a Habitat agent in Habitat Sim is precisely what AI Habitat is already designed to do. We include this operating mode to demonstrate that
    \begin{inparaenum}[(i)]
    \item a learning-based embodied agent trained in an environment without physics can be simulated with physics, and
    \item the \verb|ros_x_habitat| interface does not introduce any performance penalty and the runtime overhead is not excessive.
    \end{inparaenum}
    % The design itself is straightforward (see Figure \ref{fig:architecture}(a) for the notation): \verb|/env_node| encapsulates a Habitat Sim v2 environment; \verb|/agent_node| encapsulates a Habitat v2 agent. At each step, \verb|/env_node| publishes visual observations and localization data to \verb|/agent_node| for it to produce a discrete action and publish it to the \verb|/action| topic.  The state of the simulation environment is updated by \verb|env_node| upon receiving the action. If physics simulation is turned on, \verb|/env_node| would convert the action to a sequence of velocity commands following Algorithm~\ref{alg:action_to_vel} before stepping.
    
    \item \textbf{A ROS-based planner (\texttt{move\_base} \cite{movebase} in our case) navigates in a Habitat Sim-rendered scene (with physics).}
    Here we replace \verb|/agent_node| with a ROS-based planner node in the control loop. During navigation, we feed visual observations and a prebuilt 2D occupancy map to the ROS planner for localization and trajectory planning. Note that
    \begin{inparaenum}[(i)]
    \item depth observations are emulated as laser readings with \verb|/depthimage_to_laserscan| \cite{depthimage-to-laserscan} before being injected into the planner.
    % \item between \verb|/depthimage_to_laserscan| and the planner we omitted several nodes and topics (such as \verb|/n_hector_mapping| and \verb|/tf|) for brevity.
    \end{inparaenum}
    
    \item \textbf{A Habitat agent navigates in a Gazebo-rendered scene.}
    % Under this mode, we replace \verb|/env_node| with a node that bridges ROS with Gazebo (\verb|/gazebo|). The \verb|/gazebo_habitat_agent_bridge| node handles the communication between Gazebo and the Habitat agent: it takes sensor readings published from Gazebo and translates them to messages in formats easy for \verb|/agent_node| to process. It also takes discrete actions from \verb|/agent_node|, converts them to velocity commands, and sends the commands to \verb|/gazebo|.
    Although not demonstrated here, the nodes we designed for this mode also make it easy to swap in other ROS-bridged simulators or physical robots because those entities typically expose a ROS interface similar to Gazebo's.
\end{compactenum}

\section{Analysis of Impact of Interface and Physics} \label{sec:hab-hab}
\begin{table*}
    \small
    \centering
    \begin{threeparttable} 
    \caption{Simulation in Habitat Sim v2 with versus without physics. Habitat Sim v2 does not currently support simulation with physics for its visual navigation agents, but \texttt{ros\_x\_habitat} allows such cross-simulation.
    }
    \label{tab:physics_vs_non-physics}
    \begin{tabular}{p{0.12\linewidth} | p{0.2\linewidth} | p{0.23\linewidth} | p{0.25\linewidth}}
        \hline
         & \textbf{Agent's Geometry} & \textbf{Agent's Physical Properties} & \textbf{Simulation of Motion} \\
        \hline
        \hline
        \textbf{Simulation without physics} & A cylinder $0.1$m in radius, $1.5$m in height. & Friction coefficient is undefined; mass defined but not used for simulation. & World state advances by one \textit{action step} for each discrete action simulated; for each action the agent is translated or rotated instantaneously; no forces simulated.\\
        \hline
        \textbf{Simulation with physics} & Defined by the 3D asset attached to agent's scene node. & Friction coefficient is defined; mass is defined and used to compute dynamics. & World state advances by one \textit{continuous step} for each velocity command; forces fully simulated.\\
        \hline
    \end{tabular}
    \end{threeparttable} 
\end{table*}

A ROS bridge for the Habitat platform would be ineffective if it cannot leverage Habitat Sim v2's physics-based simulation. While Habitat Sim v2 offers physics-based simulation, the vision-based navigation agents that come with the platform were trained without physics, and Habitat Sim v2 currently cannot simulate these agents accurately with physics because they produce discrete actions. Moreover, if upon introducing physics the bridge either significantly degrades the navigation performance of an embodied agent or leads to unacceptable runtime overhead it would still be an impractical design. We dedicate this section to answer two research questions inspired by these issues.

\textbf{(RQ1) Given a Habitat navigation agent that was trained in Habitat Sim using discrete time and without physics, how effective is its navigation in the same environments but with continuous time and physics enabled?} Having a Habitat agent's discrete actions converted to a sequence of velocity commands allows us to leverage Habitat Sim's physics engine to simulate the actuation process in a more realistic fashion.  We would like to measure how much of an impact physics has on navigation performance and execution speed.

\textbf{(RQ2) Does our implementation using ROS middleware impair navigation performance or introduce unacceptable runtime overhead?} First, we would like to verify that adding the ROS interface does not alter a Habitat agent's navigation performance within Habitat Sim. Second, we would like to measure how much ROS overhead impacts execution speed.  If the simulation throughput drops below real-time---one simulated second takes more than one second to simulate computationally---then we lose a significant benefit of simulation compared to testing in the real-world.

Think of Operating Mode (a) of \verb|ros_x_habitat| as enabling a form of regression test: The goal is to demonstrate that physics and/or the ROS interface do not break the impressive capabilities of pre-trained Habitat agents. The research questions specify our regression criteria, and the subsections below describe the experiments used to test them (Section~\ref{sec:hab-hab-setup}) and the results of those experiments (Section~\ref{sec:results_and_ana}). 

\subsection{Experiment Setup}
\label{sec:hab-hab-setup}

\textbf{Agent, task and dataset.} We evaluate the Habitat v2 RGBD agent \cite{szot2021habitat} on the PointGoal navigation task (in which the agent navigates from a pre-set initial position to a goal position \cite{savva2019habitat}). The agent was trained in Habitat Sim v2, but without physics and in discrete time.  It can output one of four discrete action commands at each timestep: \texttt{move\_forward}, \texttt{turn\_left}, \texttt{turn\_right} or \texttt{stop} \cite{savva2019habitat}. We used the 1,008 navigation episodes (each episode is an instance of the task) from the Matterport3D test set \cite{chang2017matterport3d}. Because the original Habitat RGBD agent was evaluated on the Matterport3D dataset, using it leads to more meaningful benchmarking.
% for two main reasons:
% \begin{inparaenum}[(i)]
%   \item The original Habitat RGBD agent was trained in the Matterport3D training set. Choosing this dataset therefore allows better regression and benchmarking comparison;
%   \item The Matterport3D test set is publicly available. Each navigation episode ends after the agent issues the \verb|STOP| command or the agent has taken $500$ steps.
% \end{inparaenum}

\textbf{Evaluation metrics.} We employ the Success Weighted by Path Length (\verb|spl|) metric to evaluate the RGBD agent's navigation performance in each episode~\cite{savva2019habitat, szot2021habitat}. This metric lies in the range $[0, 1]$ and measures the length of the traversed path relative to the shortest path from the source to the destination; closer to $1$ implies a shorter path and thus better performance. To capture our interface's impact on execution speed we measured the total wall clock runtime (\verb|running_time|) of evaluating an agent's navigation over all $1,008$ episodes, which accounts for overheads introduced by ROS, such as initialization and inter-process communication.

\textbf{Enabling physics and mapping from discrete to continuous action space}. In Table \ref{tab:physics_vs_non-physics} we summarize how Habitat Sim v2 operates with and without physics, where
\begin{inparaenum}[(i)]
    \item The 3D asset we attached to our agent is the \verb|LoCoBot| model provided in Habitat Sim's code base \cite{savva2019habitat};
    \item An \textit{action step} is defined as one update of the world's state due to the agent completing an action in a discrete time simulation without physics; 
    \item A \textit{continuous step} is defined as the advancement of a world's state over a predefined time interval ($\frac{1}{60}$ second in our experiments) in a continuous time simulation with physics enabled.
\end{inparaenum}
As suggested in Table \ref{tab:physics_vs_non-physics}, simulation without physics teleports the agent instantaneously to the destination position of the discrete action, whereas simulation with physics requires the agent to move through space at a specified velocity. To transfer the existing RGBD agent, or any agent that outputs discrete actions, to the continuous time simulation with physics, we use Algorithm \ref{alg:action_to_vel} to map each discrete action to a sequence of velocity commands, where
\begin{inparaenum}[(i)]
    \item $control\_period$ is a user-supplied parameter and defines the time in seconds it takes an agent to complete an action step. Set to $1$ second in our experiments;
    \item $steps\_per\_sec$ defines the number of continuous steps Habitat Sim advances for each second of simulated time. Set to $60$ in our experiments;
    \item the angular velocities are measured in degrees / second; and
    \item the linear velocities are measured in meters / second.
\end{inparaenum}

\begin{algorithm}
	\caption{Convert a discrete action from a Habitat agent to a sequence of velocities. The conversion allows the interface to invoke Habitat Sim v2's built-in API  (\texttt{step\_physics()}) to simulate the action with physics.}
	\label{alg:action_to_vel}
	\begin{algorithmic}[1]
	    \Require a discrete $action$ as one of the following: \texttt{move\_forward},  \texttt{turn\_left}, or \texttt{turn\_right}
	    \State initialize $linear\_velocity$, $angular\_velocity$ as zero vectors
	    \State {$num\_steps = control\_period \cdot steps\_per\_sec$}
	    \If {$action == \texttt{move\_forward}$} 
            \State $linear\_velocity = [0.25 / control\_period, 0, 0]$
        \ElsIf {$action == \texttt{turn\_left}$}
            \State $angular\_velocity = [0, 0, 10.0 /control\_period]$
        \ElsIf {$action == \texttt{turn\_right}$}
            \State $angular\_velocity = [0, 0, -10.0 /control\_period]$
        \EndIf
		\For {$count\_steps=1, 2, \ldots, num\_steps$}
			\State env.step\_physics($(linear\_velocity, angular\_velocity)$)
		\EndFor
	\end{algorithmic} 
\end{algorithm}

\textbf{Experiment Configurations.}
We conducted our experiments on four configurations in order to independently observe the impact of introducing physics-based simulation and adding ROS.
\begin{inparaenum}[(i)]
    \item \verb|-Physics & -ROS|. We have the Habitat RGBD agent actuate in its default, discrete action space without using the ROS middleware, and we run Habitat Sim without physics turned on. This setting is the configuration in which the agent was trained, and serves as a baseline for our experiment. 
    \item \verb|+Physics & -ROS|. We enable physics-based simulation, but do not use ROS. 
    \item \verb|-Physics & +ROS|. The Habitat RGBD agent communicates with Habitat Sim v2 through the ROS interface, as shown in Figure \ref{fig:architecture}(a). The agent still navigates using its discrete action space, and the simulation is run without physics.
    \item \verb|+Physics & +ROS|. The combination of the previous two settings.
\end{inparaenum}

% \textbf{The problem of reproducibility.} Given sensor observations as input, Habitat's reinforcement learning agents use a neural network to generate an action; however, the action is chosen non-deterministically by default: The output action is sampled from the entire action space, with each action's probability of being sampled equal to its confidence score from the last layer. Sampling is implemented with a standard pseudo-random number generator, so we can impose determinism and achieve reproducible results by fixing the random number generator's initial seed.

% In order to build confidence that the choice of initial seed does not impact experimental outcomes too significantly, we ran the \verb|-Physics & -ROS| configuration (the fastest, since it does not include physics or ROS) using ten randomly chosen seeds.  The variability in navigation and timing metrics between seeds was dwarfed by the variability between scenarios, so for the remaining (slower) configurations we ran only a single seed.

\textbf{Evaluation platform.} We ran all experiments on a desktop with an i7-10700K CPU, 64 GB of RAM, and an NVIDIA RTX 3070 GPU under Ubuntu 20.04, ROS Noetic and Habitat v0.2.0.

\subsection{Results and Analysis} \label{sec:results_and_ana}
In this section we present the results from our experiments and seek to answer the two research questions posed earlier.
 
    \begin{figure}
        \centering
        \includegraphics[width=.45\textwidth]{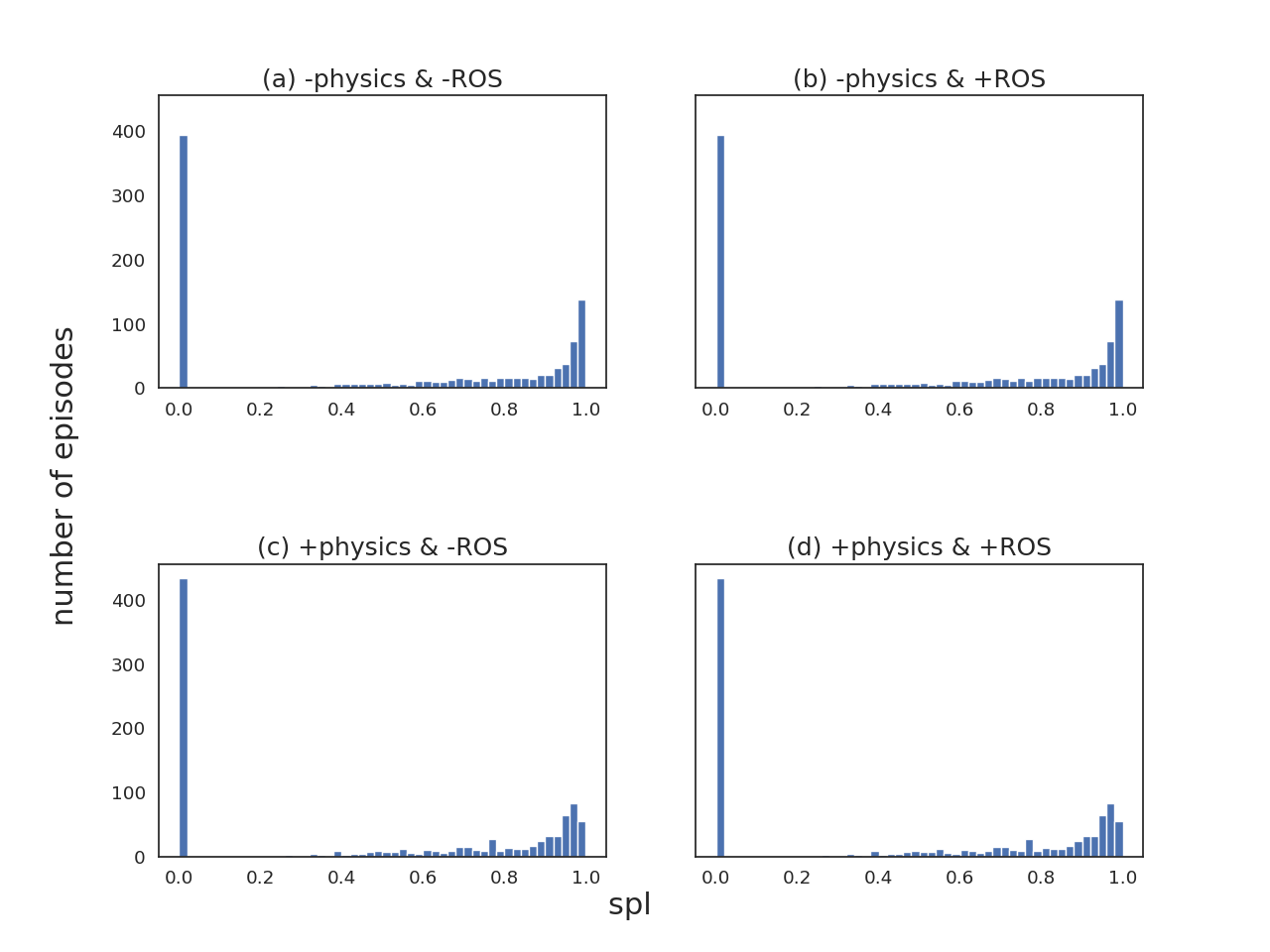}
        \caption{Distribution of \texttt{spl} under the four experimental configurations. Failed episodes have \texttt{spl} at $0$. Under Configurations (a) and (b), the mean \texttt{spl} produced by the v2 RGBD agent is $0.495$ without physics, which is close to the mean \texttt{spl} ($0.53$) reported for the v1 RGBD agent in \cite{savva2019habitat}. Under Configurations (c) and (d), the \texttt{spl} dropped slightly to $0.455$ after physics was enabled.}
        \label{fig:spl_4_configurations}
    \end{figure}
    
    \begin{figure}
        \centering
        \includegraphics[width=.45\textwidth]{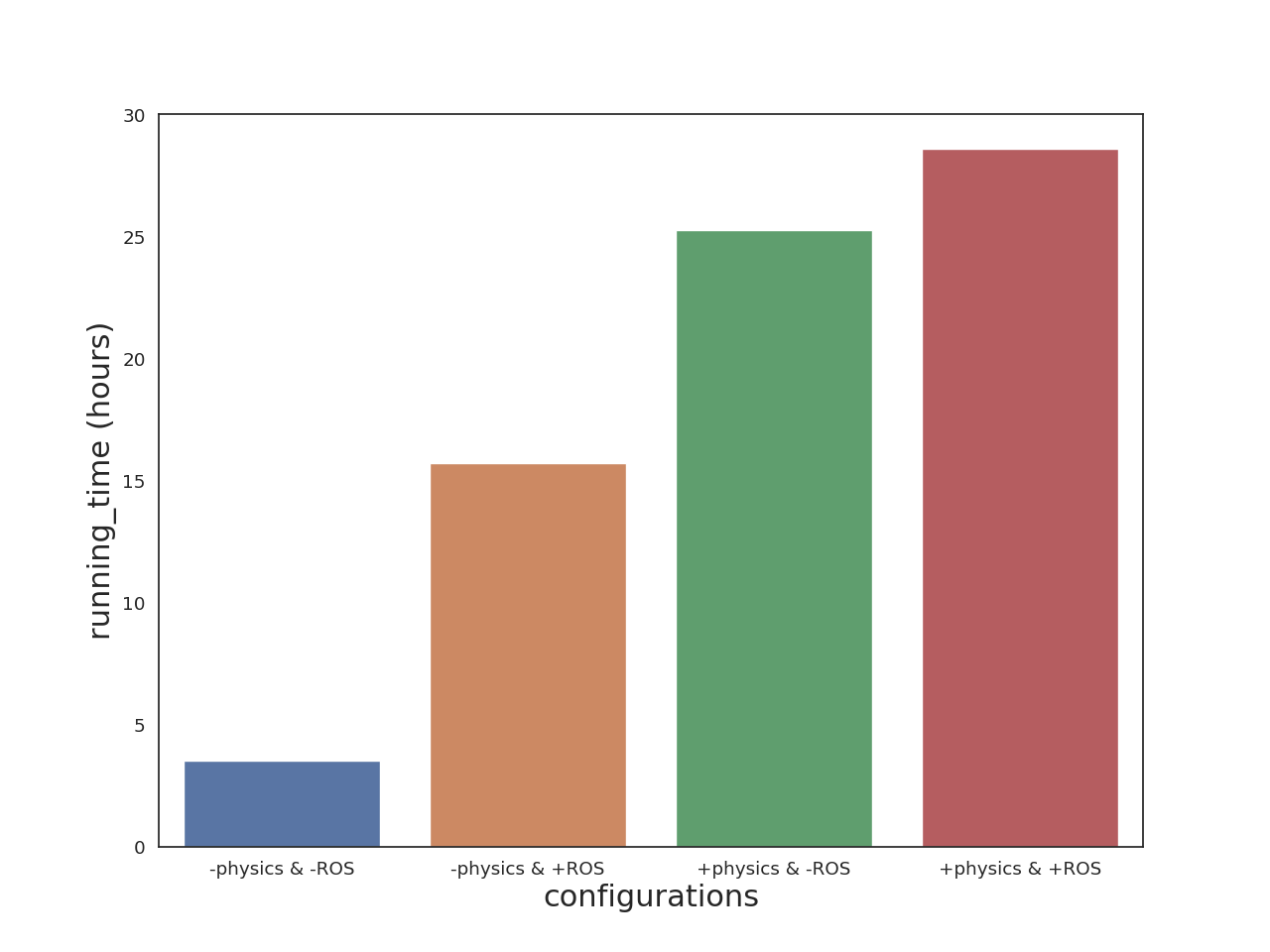}
        \caption{Total runtime under the four experimental configurations for all $1,008$ episodes. 
        %Under configuration \texttt{+Physics \& +ROS|}, our interface can still achieve real-time performance: each action step (which lasts $1$ real-world second) can be simulated in $0.4$ s.
        }
        \label{fig:running_time_4_configurations}
    \end{figure}

\textbf{(RQ1) Given a Habitat navigation agent that was trained in Habitat Sim using discrete time and without physics, how effective is its navigation in the same environments but with continuous time and physics enabled?}
Comparing subplots (a) and (c) in Figure \ref{fig:spl_4_configurations} we can see the effect that enabling physics has on the distribution of \verb|spl| over the 1,008 episodes. We see that the number of failed episodes (those with $\verb|spl| = 0.0$) increased (by 129) after we mapped actions to velocities and introduced physics-based simulation, but the average \verb|spl| only dropped slightly (from $0.495$ to $0.455$). 
We visually examined the 129 episodes in which the agent navigated successfully in the discrete action space but failed in the continuous action space. For the majority of them (80/129) the failure was due to the robot being ``stuck'', i.e. the agent is not able to move forward but remains within the scene. In some episodes the agent was stuck because the agent's movement is constrained to two dimensions by our implementation of Algorithm~\ref{alg:action_to_vel}: It was not able to climb stairs when \verb|move_forward| is converted to a sequence of planar velocities. This problem does not manifest in the discrete time mode without physics: it appears that the agent is automatically teleported to the correct height for whatever planar position it should occupy.
%Four causes of failure came to our attention:
% \begin{inparaenum}[(i)]
%     \item Mesh cavities. This term refers to empty regions in a Matterport3D scene in which mesh vertices and surfaces are undefined. Because the agent visits many more states when physics is turned on---it must traverse intermediate states during any motion, rather than just jumping to the final state---it is more likely that an agent will blunder into one of these cavities.  We found 15/129 failed cases in which the agent moved into a mesh cavity and was not able to get out.
%     \item 
%     \item The agent issued the \verb|STOP| action at a distance close to the goal but slightly larger than $0.2$m; consequently, the evaluation routine considered it a failed navigation (9/129).
%     \item The agent chose a poor path and was unable to achieve the goal within $500$ steps (25/129).
% \end{inparaenum}

Figure \ref{fig:running_time_4_configurations} shows our timing result. Adding just physics (without ROS) slows Habitat by a factor of roughly eight, as extra time was required for simulating a discrete action as a sequence of continuous actions (Algorithm \ref{alg:action_to_vel}), and for simulator reset (to load and delete physics-based object assets and reconfigure the simulator).
    
We conclude that using physics will significantly reduce the throughput of the simulation engine, but the modest performance degradation confirms that training RL agents in the simplistic but high-throughput discrete action configuration without physics can produce reasonable results even for agents intended to be used in more realistic environments~\cite{discrete-drl}, and the full physics need not be turned on until validation or even final testing.

\textbf{(RQ2) Does our implementation using ROS middleware impair navigation performance or introduce unacceptable runtime overhead?}  In terms of navigation performance, once we removed non-determinism from the system we were able to show that the introduction of ROS did not affect navigation performance regardless of whether physics was enabled: Each episode had the exact same number of steps with and without ROS, and the SPL discrepancies were at the level of floating point round-off.  Comparing the distributions in the left and right columns of figure~\ref{fig:spl_4_configurations} shows the latter effect qualitatively.

Turning to execution time: Figure \ref{fig:running_time_4_configurations} shows that adding ROS increases total runtime by roughly a factor of five (without physics) and less than 20\% (with physics).  The increases are due to the overheads from inter-process communication between ROS nodes, as well as intra-process communication between threads running various service handlers and subscriber callbacks within each node. Although the increase in running time is significant, it is acceptable: with overhead from both ROS and physics taken into account, simulating over $274,000$ action steps across the full set of episodes in less than $30$ hours implies an average computation time of less than $0.4$ seconds for each action step lasting one second of simulated time; consequently, the throughput is more than twice real-time.

\section{Navigation of ROS-based Planners in Habitat Sim} \label{sec:ros-hab}
\begin{figure}
    \centering
    \includegraphics[width=.45\textwidth]{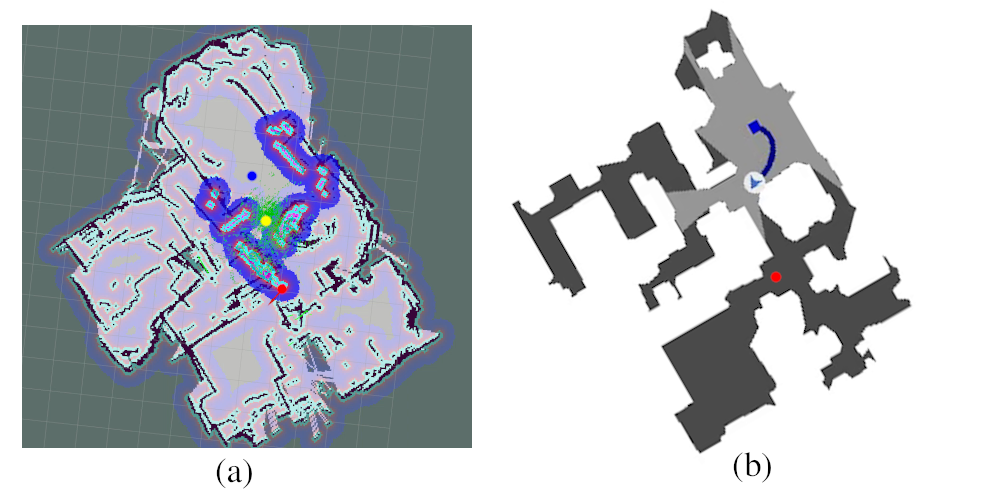}
    \caption{ROS-based planner \texttt{move\_base} navigating in Matterport3D scene \texttt{2t7WUuJeko7} simulated by Habitat Sim v2. (a) The agent's final position overlayed on top of the map build by \texttt{rtab\_map\_ros}. (b) Top-down map of the scene from Habitat Sim v2. The blue curve indicates the agent's trajectory. Blue square: starting position of the agent. Yellow dot: final position. Red dot: goal position.}
    \label{fig:planner_in_hab_2t7}
\end{figure}
\begin{figure}
    \centering
    \includegraphics[width=.45\textwidth]{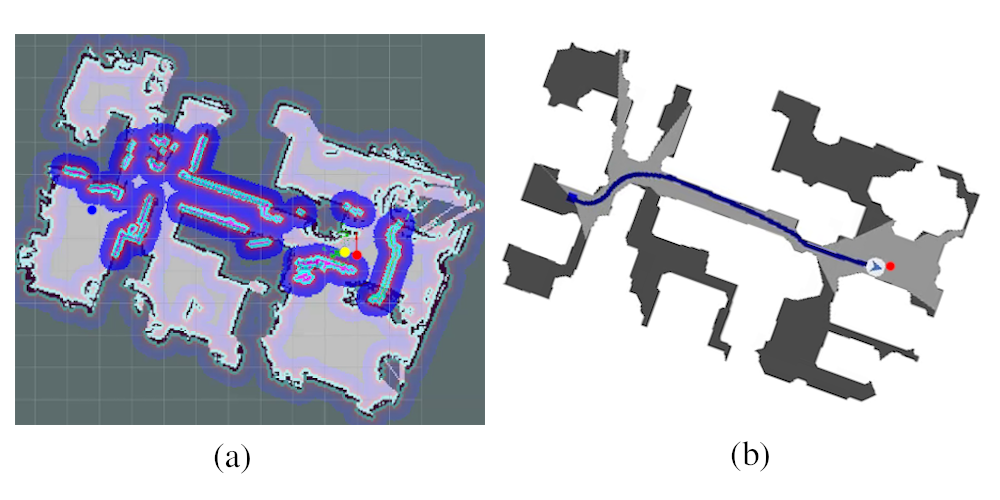}
    \caption{ROS-based planner \texttt{move\_base} navigating in Matterport3D scene \texttt{RPmz2sHmrrY} simulated by Habitat Sim. (a) The agent's final position and laser-scanned map. (b) Top-down map from Habitat Sim.}
    \label{fig:planner_in_hab_RPm}
\end{figure}

For robotics researchers interested in designing and testing navigation algorithms---be they classical or RL---we believe that this configuration of \verb|ros_x_habitat| will be most useful, as Habitat Sim provides a novel, high throughput, photorealistic simulation environment in which to test such algorithms.  In this section we do not seek to show the merits of a particular mapping, planning or navigation system, but simply that some standard packages from ROS can be easily and successfully connected to Habitat Sim in the configuration shown in Figure~\ref{fig:architecture}(b).

First, we mapped two scenes from the Matterport3D dataset with \verb|rtab_map_ros| \cite{rtabmap}. Second, we attached the \verb|LoCoBot| \cite{savva2019habitat} asset to the agent in Habitat Sim (the same asset used in the previous section's experiments).  Finally, we manually set a goal position for the \verb|move_base| \cite{movebase} planner.  Figures \ref{fig:planner_in_hab_2t7} and \ref{fig:planner_in_hab_RPm} show the generated map (left) and ground-truth map (right) overlaid with the final agent position and sensor readings (left) and path (right) for the two scenes.  The planner failed to reach the goal in Figure~\ref{fig:planner_in_hab_2t7} but succeeded in Figure~\ref{fig:planner_in_hab_RPm}.

We observed during the simulations that the agent often had a hard time localizing itself, especially in cluttered regions.  We expect that with some tuning of the mapping, planning and navigation algorithms' parameters we could achieve better performance.

\section{Navigation of Habitat Agents in Gazebo} \label{sec:hab-gazebo}
For RL researchers interested in testing their agents on real robots, this configuration demonstrates the benefits of \verb|ros_x_habitat|.  Researchers familiar with ROS know that despite the significant sim2real gap in the Gazebo simulator, it can be an effective first target during design and testing because it will expose whole classes of common design bugs, such as incorrectly typed or connected data flows, coordinate transform errors, and gross timing issues.

% \begin{figure}
%     \centering
%     \includegraphics[width=.45\textwidth]{figures/turtlebot3_house-3d_and_map.png}
%     \caption{The \texttt{House} scene from ROS package \texttt{turtlebot3\_gazebo}. a) a view of the scene in 3D; b) a map scanned from the scene using \texttt{hector\_mapping} \cite{hector}.}
%     \label{fig:turtlebot3_house}
% \end{figure}

Although we do not demonstrate a Habitat agent connected to a physical robot in this paper, by demonstrating a Habitat agent connected to Gazebo we show that \verb|ros_x_habitat| can allow connection of Habitat agents to other simulation environments that have ROS bridges.  Note that we do not expect the Habitat agent to perform particularly well in this Gazebo environment since it was trained in the much richer visual and geometric environments of Matterport3D in Habitat Sim.

\begin{figure}
    \centering
    \includegraphics[width=.48\textwidth]{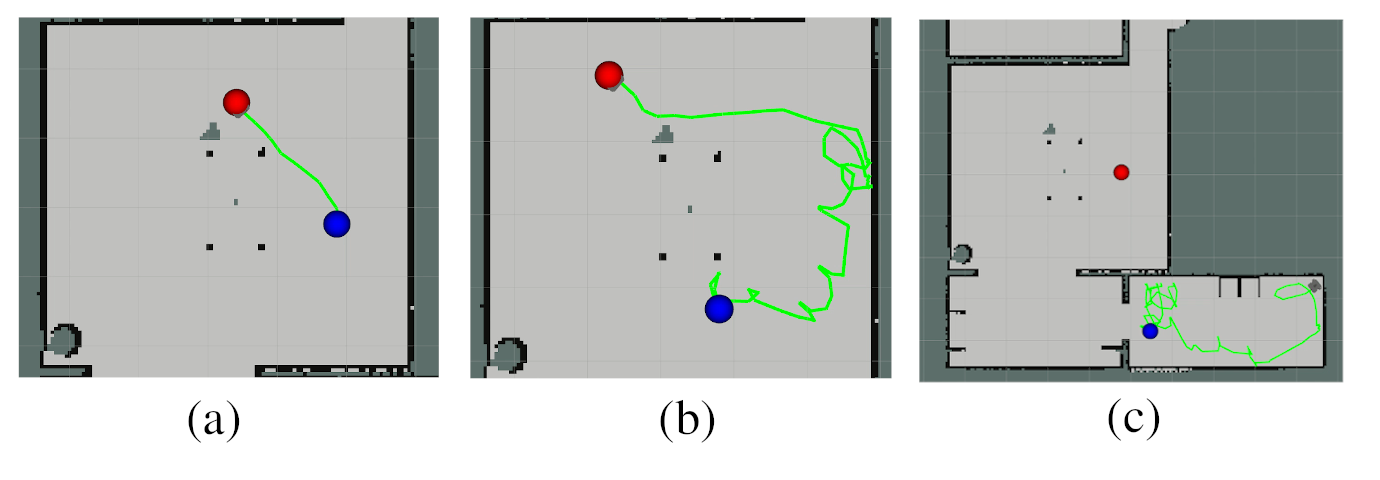}
    \caption{Maps showing the Habitat v2 RGBD agent navigating in three episodes in the \texttt{House} scene. The blue spheres indicate the starting positions of each episode; the red spheres indicate the goal positions of each episode. The green curves represent paths traversed by the agent.}
    \label{fig:hab_agent_in_gazebo}
\end{figure}

Injecting the Matterport3D dataset into Gazebo is not practical, so we used the \verb|House| scene from the \verb|turtlebot3_gazebo| package \cite{turtlebot3-gazebo} to build three navigation episodes of increasing path length as shown in Figure \ref{fig:hab_agent_in_gazebo}. Then we instantiate a Habitat v2 RGBD agent inside each episode, and let it navigate until it reaches the goal or has generated $500$ actions. Since we only intend to show that this configuration works, we do not report quantitative metrics such as \verb|spl| for these episodes.  We see that the agent succeeded in the two shorter of the three episodes, although in the middle episode the agent took a path much longer than would be optimal.

\section{Conclusion and Future Work}
We presented an extensible interface that bridges the AI Habitat platform with ROS middleware. Through this interface, researchers can develop and test their embodied agents in both established simulators such as Gazebo or the state-of-the-art Habitat Sim v2. Moreover, our interface implements the necessary transformations to allow agents trained in the original, high-throughput discrete time Habitat Sim environment without physics to navigate in the more realistic continuous time environment with physics available in Habitat Sim v2. Using the SPL metric, we show that our interface introduces negligible impacts on the navigation performance of embodied agents in a discrete action space, and we show that if the agents are actuated in a continuous action space with physics-based simulation enabled, the run time overhead is still acceptable. Although we only include two simulator options in our current work, our modular design allows easy integration of additional robotic embodiments, such as other simulators or physical robots. 

The interface is still in its infancy. Some planned work for the future of \verb|ros_x_habitat| include:
\begin{inparaenum}[(i)]
    \item Allowing the simulation of embodied agents trained for tasks other than navigation, such as the object picking task introduced in Habitat v2 \cite{szot2021habitat}.
    \item Allowing classical ROS planners to be evaluated on Habitat's navigation metrics, such as \texttt{spl} when simulated under Habitat Sim v2.
    \item Deploying an embodied Habitat agent on physical robot(s).
\end{inparaenum}


\section*{Acknowledgment}
The authors were partially supported by National Science and Engineering
Research Council of Canada (NSERC) Discovery Grant \#RGPIN-2017-04543. The authors would also like to thank Bruce Cui from the Department of Mechanical Engineering at UBC for his initial work on \verb|ros_x_habitat|, and the AI Habitat team from Facebook AI Research, including Prof. Dhruv Batra, Alex Clegg, Prof. Manolis Savva, and Erik Wijmans for their incredible support throughout our development of the interface.



%\bibliographystyle{IEEEtran}
%\begin{thebibliography}{1}

%H.~Kopka and P.~W. Daly, \emph{A Guide to \LaTeX}, %3rd~ed.\hskip 1em plus
  %0.5em minus 0.4em\relax Harlow, England: Addison-Wesley, 1999.

%\end{thebibliography}

{\bibliographystyle{./IEEEtran} % use IEEEtran.bst style
\bibliography{IEEEexample}
}

\end{document}